\documentclass[10pt, a4paper, twocolumn]{article} 
\usepackage[margin=0.75in]{geometry} 

\usepackage{cite}
\usepackage{amsmath,amssymb,amsfonts}
\usepackage{algorithmic}
\usepackage{graphicx}
\usepackage{textcomp}
\usepackage{xcolor}

\usepackage[font=footnotesize]{subcaption}
\usepackage[font=footnotesize]{caption}

\usepackage{url}
\usepackage{xurl} 

\usepackage{hyperref} 
\hypersetup{pdfborder={0 0 0}} 
\usepackage{orcidlink} 
\usepackage{authblk} 

\makeatletter
\let\oldthebibliography\thebibliography
\renewcommand{\thebibliography}[1]{%
  \oldthebibliography{#1}%
  \small%
  \setlength{\itemsep}{0pt}%
  \setlength{\parskip}{0pt}%
}
\makeatother

\begin{document}

\title{SEG-JPEG: Simple Visual Semantic Communications for Remote Operation of Automated Vehicles over Unreliable Wireless Networks}

\newcommand{\fundingnote}{This work has been supported by the GREEN-LOG project that has received funding from the European Union's Horizon Europe research and innovation programme under Grant Agreement No 101069892 and from the United Kingdom Research and Innovation (UKRI) agency. Views and opinions expressed are however those of the author(s) only and do not necessarily reflect those of the European Union, the European Climate, Infrastructure and Environment Executive Agency (CINEA) or the UKRI. Neither the European Union nor the granting authorities can be held responsible for them.}

\author[1]{Sebastian Donnelly \orcidlink{0009-0009-9716-3503}}
\author[4]{Ruth Anderson \orcidlink{0000-0002-0217-4333}}
\author[5]{George Economides \orcidlink{0000-0003-0903-2458}}
\author[2]{James Broughton \orcidlink{0000-0002-9136-1586}}
\author[2]{Peter Ball \orcidlink{0000-0003-2483-0415}}
\author[3]{Alexander Rast \orcidlink{0000-0001-9934-7191}}
\author[1]{Andrew Bradley \orcidlink{0000-0001-7053-804X}}

\affil[1]{Autonomous Driving and Intelligent Transport Group, Oxford Brookes University, Oxford, UK}
\affil[2]{School of Engineering, Computing \& Mathematics, Oxford Brookes University, Oxford, UK}
\affil[3]{Artificial Intelligence, Data Analysis and Systems Institute, Oxford Brookes University, Oxford, UK}
\affil[4]{Oxfordshire County Council, Oxford, UK}
\affil[5]{Department for Transport, London, UK}

\affil[]{Corresponding author: s.donnelly@brookes.ac.uk}

\date{}

\maketitle
\begingroup
  \renewcommand{\thefootnote}{}\footnotetext{\fundingnote}
\endgroup
\setcounter{footnote}{0}

\begin{abstract}
Remote Operation is touted as being key to the rapid deployment of automated vehicles. Streaming imagery to control connected vehicles remotely currently requires a reliable, high throughput network connection, which can be limited in real-world remote operation deployments relying on public network infrastructure. This paper investigates how the application of computer vision assisted semantic communication can be used to circumvent data loss and corruption associated with traditional image compression techniques. By encoding the segmentations of detected road users into colour coded highlights within low resolution greyscale imagery, the required data rate can be reduced by 50\% compared with conventional techniques, while maintaining visual clarity. This enables a median glass-to-glass latency of below 200 ms even when the network data rate is below 500 kbit/s, while clearly outlining salient road users to enhance situational awareness of the remote operator. The approach is demonstrated in an area of variable 4G mobile connectivity using an automated last-mile delivery vehicle. Results indicate that large-scale deployment of remotely operated automated vehicles could be possible even on the often constrained public 4G/5G mobile network, providing the potential to expedite the nationwide roll-out of automated vehicles.

\end{abstract}


\section{Introduction}

\begin{figure}[!tp]
    \centering
    \includegraphics[width=1\linewidth]{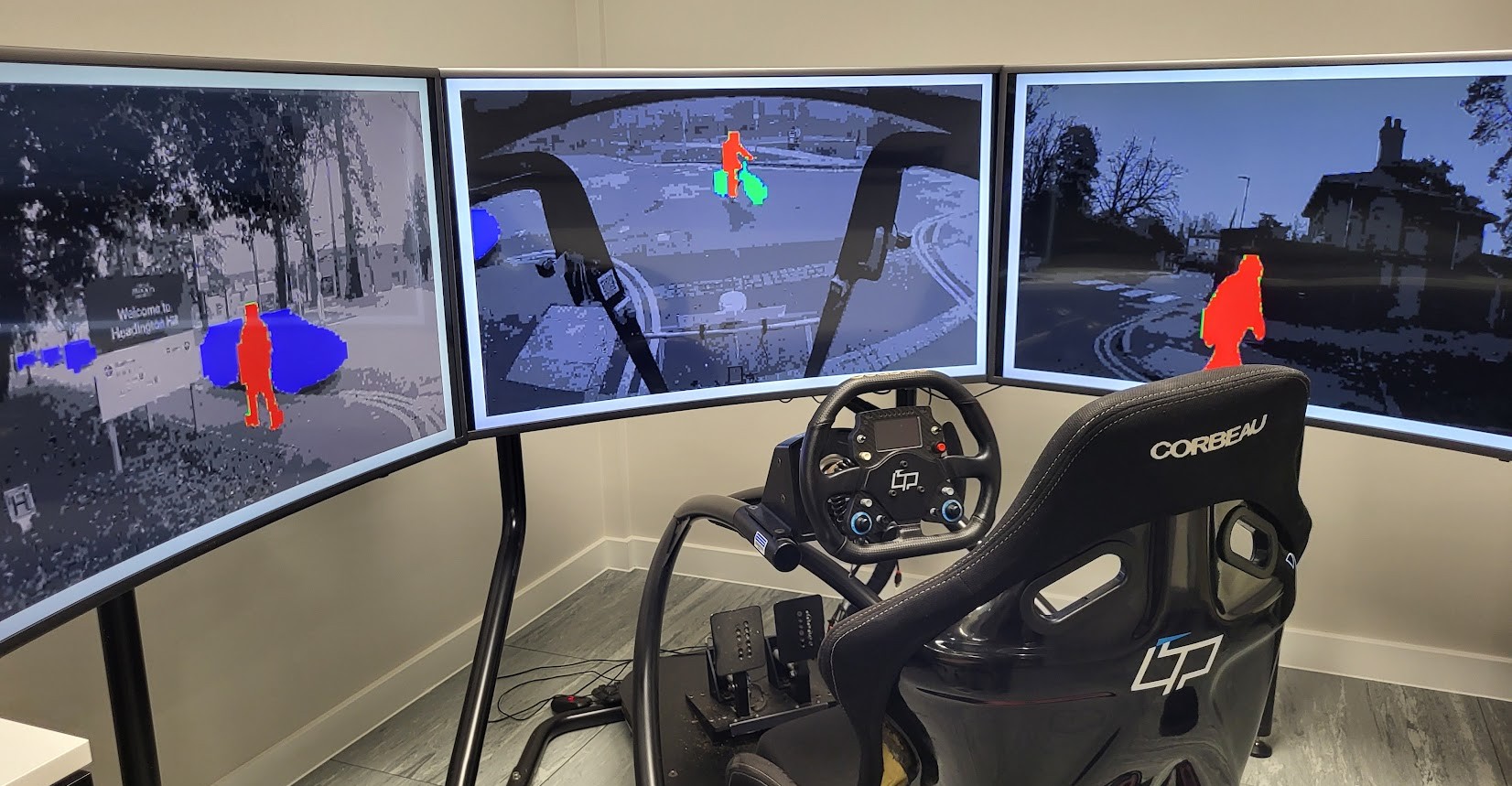}
    \caption{Augmented greyscale imagery presented to a remote operator, highlighting vehicles (blue), pedestrians (red) and \textcolor{black}{bicycles (green)}.}
    \label{fig:SegJPEGInAction}
\end{figure}

 The capability to remotely operate a vehicle from anywhere in the world represents a key step towards wider adoption of connected and automated vehicles \cite{rosa-garciaBridgingRemoteOperations2025}. This enables remote assistance to be provided to automated vehicles which, if lower than \textcolor{black}{the Society of Automotive Engineers (SAE)
level 5 categorisation}, may still require corrective input when presented with unknown challenges \cite{saeinternationalTaxonomyDefinitionsTerms2021}. To enable a Remote Operator (RO) to make safety critical decisions, a sufficient amount of information needs to be streamed at low latency from the connected vehicle for the operator to maintain situational awareness (Fig. \ref{fig:SegJPEGInAction}).

Current remote operation deployments are typically limited to geographically constrained areas, as conventional real-time imagery streaming techniques require a continuous, low-latency, high-throughput network connection - which is not guaranteed across public mobile networks \cite{amadorSurveyRemoteOperation2022}. The arrival of 6G may mitigate some of these issues, but fundamental bandwidth and range limitations, especially in the higher spectral bands, means 6G is unlikely to provide a total solution. Mobile network operators often prioritise downlink performance over uplink performance \cite{fezeuTeleoperatingAutonomousVehicles2026}. This creates a bottleneck, as streaming imagery and other sensor data from a vehicle to a RO relies heavily on a strong uplink connection. Reduced data rates typically result in compression artefacts, which can make it difficult for a RO to accurately perceive salient information, with potentially safety-critical consequences \textit{e.g.} failure to notice a pedestrian. Throughput over a network closely ties with its latency characteristics \cite{caoIndustrialMetaverseAge2023}, which has a detrimental impact on remote operation performance, as it becomes increasingly difficult to control a vehicle when the closed-loop latency is above 300 ms \cite{neumeierMeasuringFeasibilityTeleoperated2019}. Despite these challenges, the ability to remotely operate a vehicle on a relatively low throughput network is vital for the widespread adoption of Remotely Operated Vehicles (ROVs) - requiring a technique which can exceed the capability of currently available imagery compression algorithms while preserving saliency \cite{gimenez-guzmanSemanticV2XCommunications2024}.

Advances in computer vision have improved the accuracy of scene recognition \cite{dolatyabiDeepLearningTraffic2025} (\textit{e.g.} detection of objects, pedestrians, road users \textit{etc.}), presenting an opportunity for a technique which utilises both the efficiencies of modern compression algorithms and the environmental awareness enabled by computer vision. Sending only the information that is valuable to a recipient is a key concept of semantic communication - representing a paradigm shift away from more conventional communication approaches which simply send compressed bit level data \cite{huangSemanticCommunicationsDeep2023}. Such an approach could reduce the loss of situational awareness associated with traditional high compression video streaming, enabling sustained, robust communication on low throughput networks. In this work, traditional JPEG compression is enhanced by semantic encodings to overcome the limitations of highly compressed imagery, for a real-time remote control application (Fig.~\ref{fig:SegJPEGInAction}).

\section{Related Work}

A major concern of ROV deployments is the insufficient coverage of high quality mobile networks, affecting throughput and latency \cite{neumeierMeasuringFeasibilityTeleoperated2019},\cite{bokernAdvancingRemoteControl2025}. 6G rollouts are likely to follow the same pattern as previous 4G and 5G deployments - \textit{i.e.} 
for some years full 6G coverage is likely to be intermittent except in the most dense and well-connected urban areas (where heavy usage may become a factor). While Georg \textit{et~al.} \cite{georgSensorActuatorLatency2020} showed that lowering imagery compression quality can decrease network throughput requirements and latency, the associated loss of image fidelity can impact a RO’s ability to identify the information necessary to safely control a vehicle, with Hoffmann \textit{et~al.} \cite{hoffmannQuantifyingInfluenceImage2022} finding that poor quality imagery can negatively affect the RO's reaction time, thus increasing closed-loop latency.

\begin{figure} [ht]
     \centering
     \begin{subfigure}[b]{0.48\columnwidth}
         \centering
         \includegraphics[width=\textwidth]{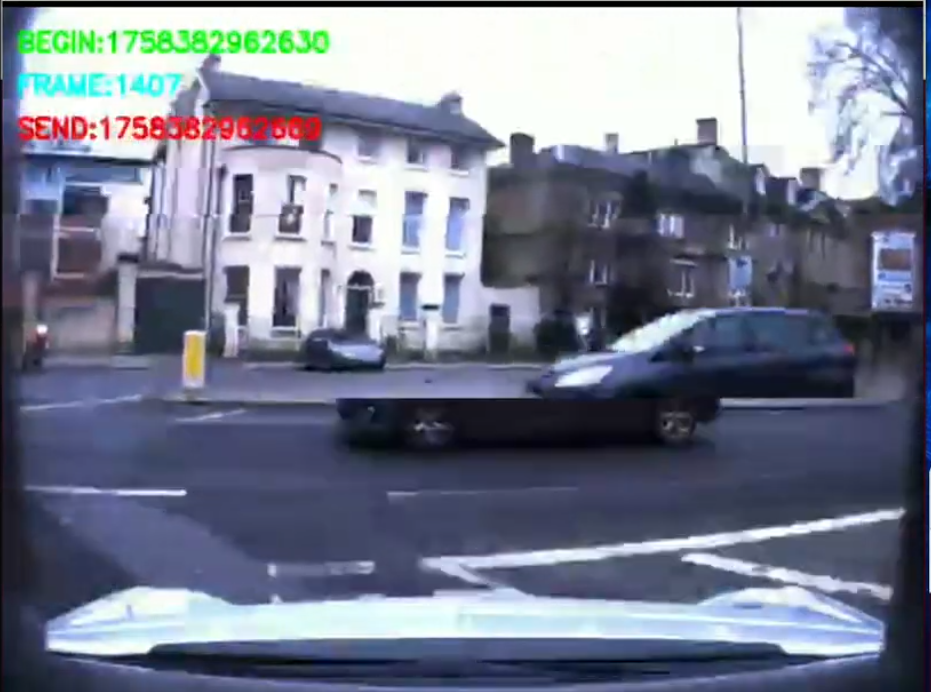}
         \caption{Frame tearing.}
         \label{fig:DistortedH264}
     \end{subfigure}
     \hfill
     \begin{subfigure}[b]{0.48\columnwidth}
         \centering
         \includegraphics[width=\textwidth]{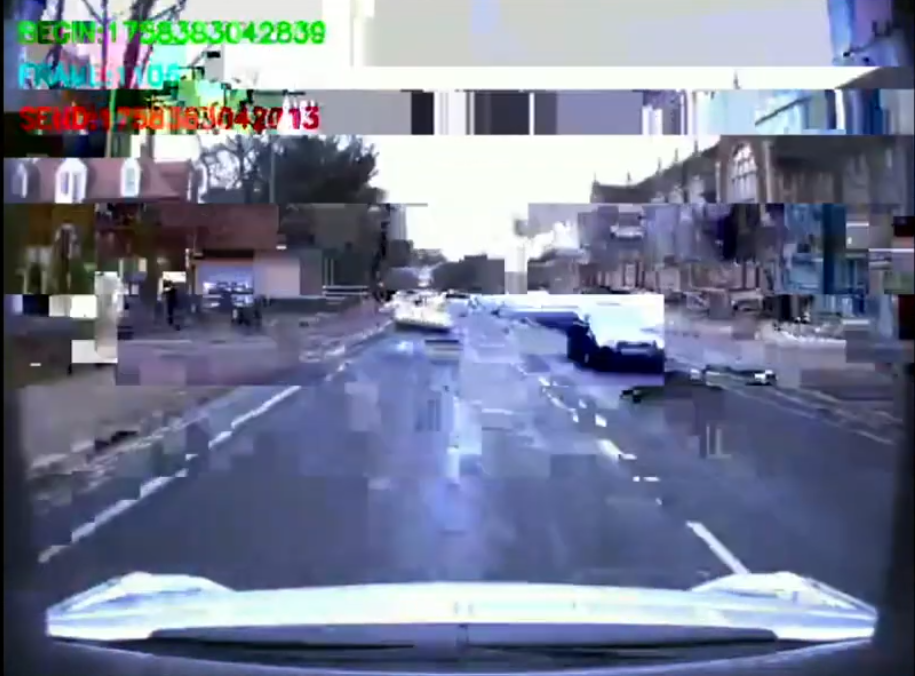}
         \caption{Macroblocking.}
         \label{fig:DistortedH264-2}
     \end{subfigure}
     \caption{Frame distortion in h.264 imagery caused by a reduced quality mobile network connection\protect\footnotemark[1].}
\end{figure}

Many recent remote driving trials use the h.264 codec to compress transmitted imagery \cite{bellangerChallengesRemoteDriving2024},\cite{kerblTUMTeleoperationOpen2025}, \cite{testouri5GEnabledTeleoperatedDriving2025}, typically requiring a throughput of at least 20 Mbit/s to provide a RO with imagery of sufficient quality to safely control a vehicle. The h.264 codec is likely the most widely used compression method across \textit{all} streaming imagery applications - achieving a high compression efficiency by using interframe compression, which involves processing of frames for the extraction of motion vectors relative to occasional keyframes prior to transmission. Originally designed for streaming services where 2 s or more latency is acceptable \cite{wiegandOverviewH264AVC2003}, \textcolor{black}{the compression efficiency of h.264 is decreased when optimised for low-latency, due to the need for regular keyframe updates.} On a variable network connection, partial motion vector updates can cause distortion such as a frame tearing or macroblocking, rendering the frame corrupt and potentially unsuitable for use by a RO (Fig.~\ref{fig:DistortedH264}). This significantly impacts upon performance for a ROV operating on a throughput-constrained mobile network.

\footnotetext[1]{Evaluated using imagery from the Oxford RobotCar dataset \cite{maddern1Year10002017}}

\textcolor{black}{The maximum error-free data rate achievable over a network connection is subject to the Shannon limit \cite{shannonMathematicalTheoryCommunication1948}. As a result, when network quality is reduced, aggressive compression is needed to maintain real-time, low latency imagery. According to rate-distortion theory \cite{shannonCodingTheoremsDiscrete1959}, reducing the bit-rate of lossy compression techniques leads to a reduction in image fidelity, introducing visual artifacts which can render the imagery unusable for a RO}. Semantic communication has the potential to increase the efficiency of data transmission by communicating meaning instead of simple uninterpreted data \cite{liangGenerativeAIDrivenSemantic2025}. While recent work on semantic communication has shown promising results in compressing imagery more efficiently than traditional codecs \cite{gimenez-guzmanSemanticV2XCommunications2024}, the focus has been on accurate image reconstruction \cite{huangSemanticCommunicationsDeep2023} rather than low latency capabilities or computational efficiency \cite{chenLLMBasedSemanticCommunication2025} - factors which are critical to remote operation systems running on computationally constrained edge hardware. 

The ability to combine the scene understanding capabilities offered by computer vision techniques with a conventional imagery compression technique would enable a semantic communication approach - conveying the information which matters to the RO in an efficient manner designed specifically for variable network environments.

\section{Methodology}

The methodology covers the development of a semantic layering imagery transmission technique, the setup of a ROV and the experimental evaluation using a public mobile network. 

\subsection{SEG-JPEG: Semantic Layering Imagery Technique}
\begin{figure} [ht]
    \centering
    \includegraphics[width=1\linewidth]{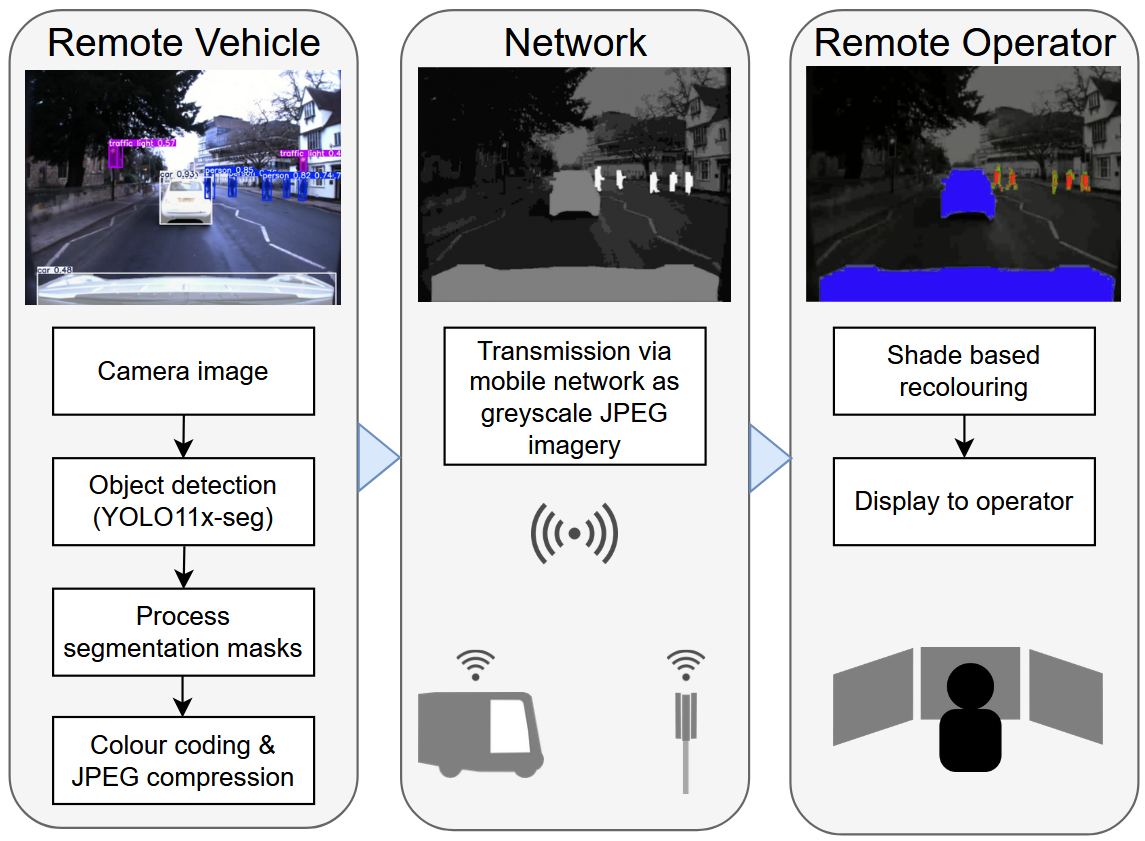}
    \caption{The SEG-JPEG approach to augment greyscale imagery\protect\footnotemark[1].}
    \label{fig:Methodology}
\end{figure}

Ensuring that an RO can understand the environment surrounding the vehicle is of primary importance to enable safe operation, and the artefacts introduced via traditional imagery compression approaches can make it difficult for a RO to clearly distinguish salient objects. The approach presented in this study (Fig.~\ref{fig:Methodology}) processes the imagery using computer vision prior to compression, augmenting low quality imagery with segmentations - clearly highlighting salient objects (\textit{e.g.} other vehicles, pedestrians \textit{etc.}). The imagery can then be transmitted to the RO in the compact greyscale format before being recoloured using reserved shades.

\subsubsection{Object Detection}

To detect and highlight road users in the imagery, the system utilises the YOLO computer vision model \cite{redmonYouOnlyLook2016}. The YOLO11x-seg model provides the highest accuracy computer vision inference of the YOLO11 family, and is capable of segmenting imagery according to the detected classifications. Using the vehicle's Nvidia RTX 4070 GPU, the average time to extract a segmented frame using the YOLO11x-seg model is 35.8 ms with a standard deviation of 3.1 ms. Although inference is much faster using the smaller YOLO11n-seg model at an average of 7.3 ms, the resulting decrease in confidence and increase in misdetections is believed to negate any benefits experienced by faster processing.

\subsubsection{Salient Compression}

Uncompressed greyscale imagery is three times more compact then conventional RGB imagery, using a single channel to store pixel luminance. While colour JPEG compression efficiency can be attributed to an aggressive reduction of colour information via chroma subsampling \cite{wallaceJPEGStillPicture1991}, greyscale JPEG compression further reduces imagery sizes by removing the colour channels entirely, instead solely using pixel luminance. To embed the detected road users within the highly compressed greyscale JPEG imagery, 3 distinct shade values are chosen at large intervals to represent the segmentation masks to be clearly defined during the later recolouring process, with the remaining shades constituting the rest of the image in greyscale. A shade value of 240 represents a person, 200 a bicycle and 160 a vehicle. Once the greyscale segmentation mask and greyscale image are combined, the imagery is compressed using greyscale JPEG compression.

\subsubsection{Operator Reconstruction}
\begin{figure} [ht]
     \centering
     \begin{subfigure}[b]{0.48\columnwidth}
         \centering
         \includegraphics[width=\textwidth]{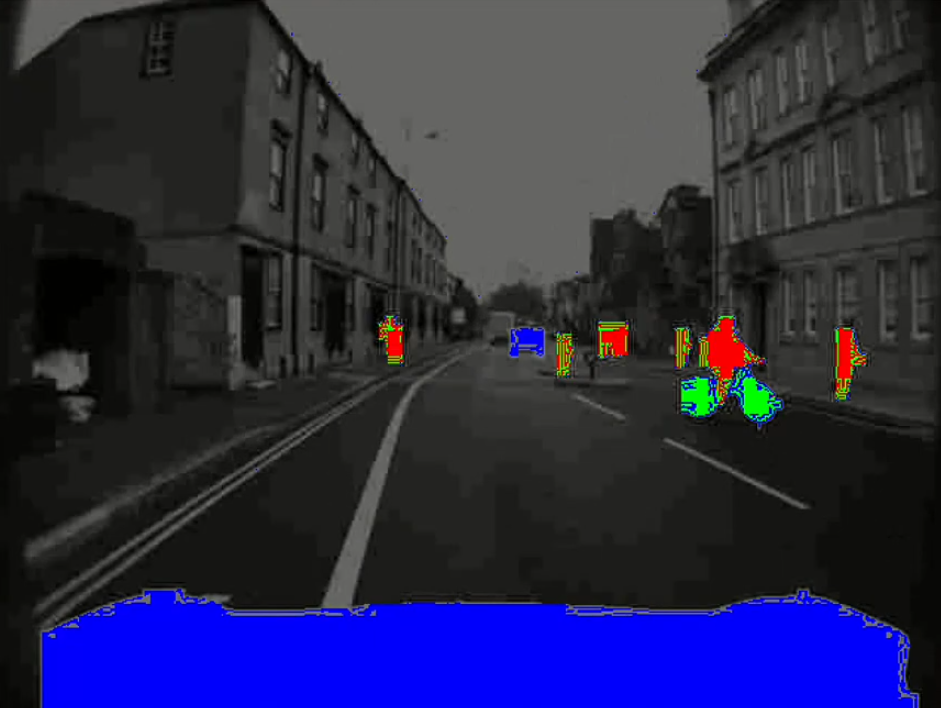}
         \caption{Busy urban road}
         \label{fig:Cyclist}
     \end{subfigure}
     \hfill
     \begin{subfigure}[b]{0.48\columnwidth}
         \centering
         \includegraphics[width=\textwidth]{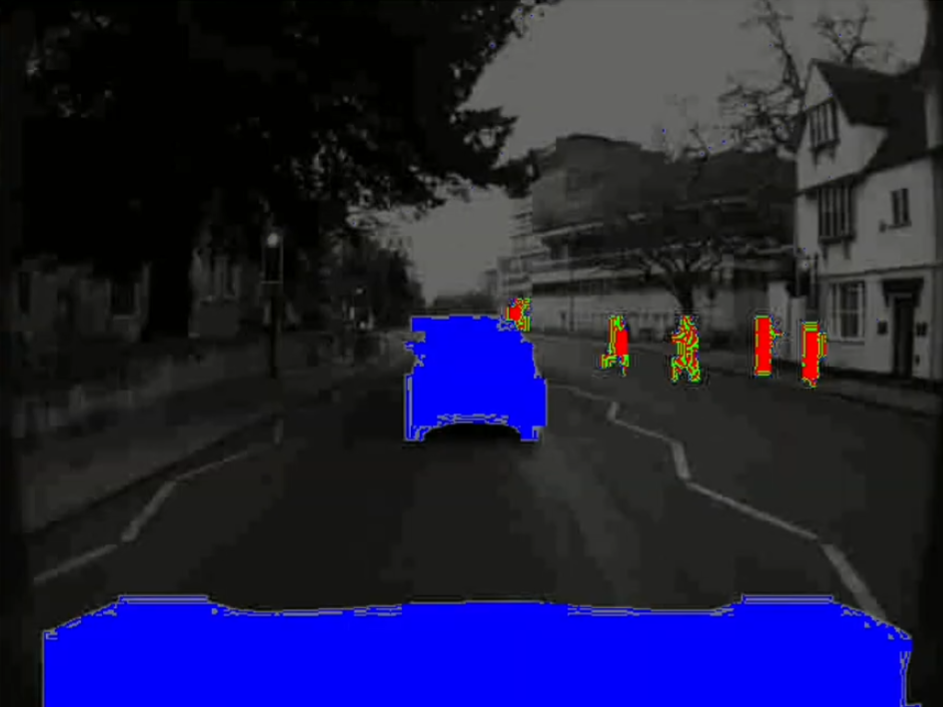}
         \caption{Pedestrians crossing road}
         \label{fig:Crossing}
     \end{subfigure}
     \caption{Highly compressed greyscale JPEG imagery with road users highlighted to enhance RO situational awareness. Red indicates a person, green a bicycle, blue a road vehicle\protect\footnotemark[1].}
\end{figure}

The received greyscale SEG-JPEG imagery is enhanced by selective recolouring according to the predefined shade values. Pedestrians, bicycles and vehicles are highlighted in primary colours, ensuring that they remain visible to the RO even if the imagery is highly compressed (Fig.~\ref{fig:Cyclist}). The buffer of 40 shades between each of the three road user masks reduces the bleeding effect on the borders caused by the JPEG compression process which can affect how the shade is interpreted.

\subsection{Hardware Setup}
\begin{figure} [ht]
    \centering
    \includegraphics[width=1\linewidth]{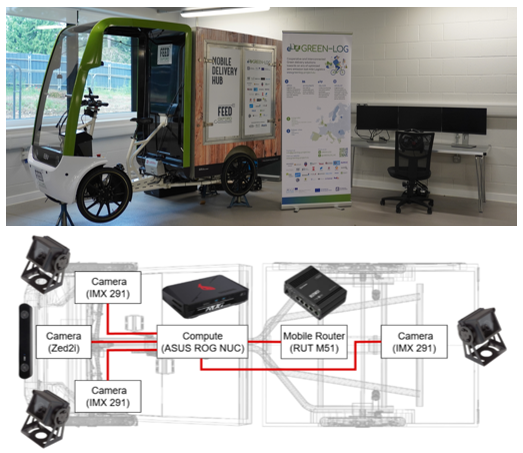}
    \caption{Green-Log automated ROV (top), remote operation hardware installed in the vehicle (bottom).}
    \label{fig:RemoteDrivingSetup}
\end{figure}

Electric Assisted Vehicles (EAVs) are gaining popularity in major cities for last-mile delivery services, including in Oxford. The Green-Log project \cite{GreenLogProject2023} equipped an EAV with actuators, sensors and computing to enable its use as a ROV, replacing the need for an in-vehicle driver (Fig.~\ref{fig:RemoteDrivingSetup}).

The remote operation system runs on a mini-PC (ASUS ROG NUC, 8 GB Nvidia RTX 4070 Mobile GPU, 32 GB RAM, Intel Core Ultra 9). Mobile communication is enabled by the RUT M51 mobile router, providing 4G/5G network access via (consumer grade) SIM card. Front facing imagery is collected by a Zed2i stereo camera, with three additional Arducam IMX 291 cameras, positioned to provide the RO with a complete view of the surrounding environment.

The interface with the vehicle is achieved via a ROS2 node running on the mini-PC - which converts steering, braking and throttle commands from the RO to CANbus commands to the vehicle. Communication between the vehicle and RO is achieved using UDP sockets via a mobile network using ZeroTier to handle routing over the Internet. Control commands, cameras and individual sensor and status messages are each directed to their own unique UDP port. Control commands and low data rate sensors use the JSON format, while higher data rate information from the cameras is sent as a bytestream.

\subsection{Experiments}

\begin{figure} [ht]
    \centering
    \includegraphics[width=1\linewidth]{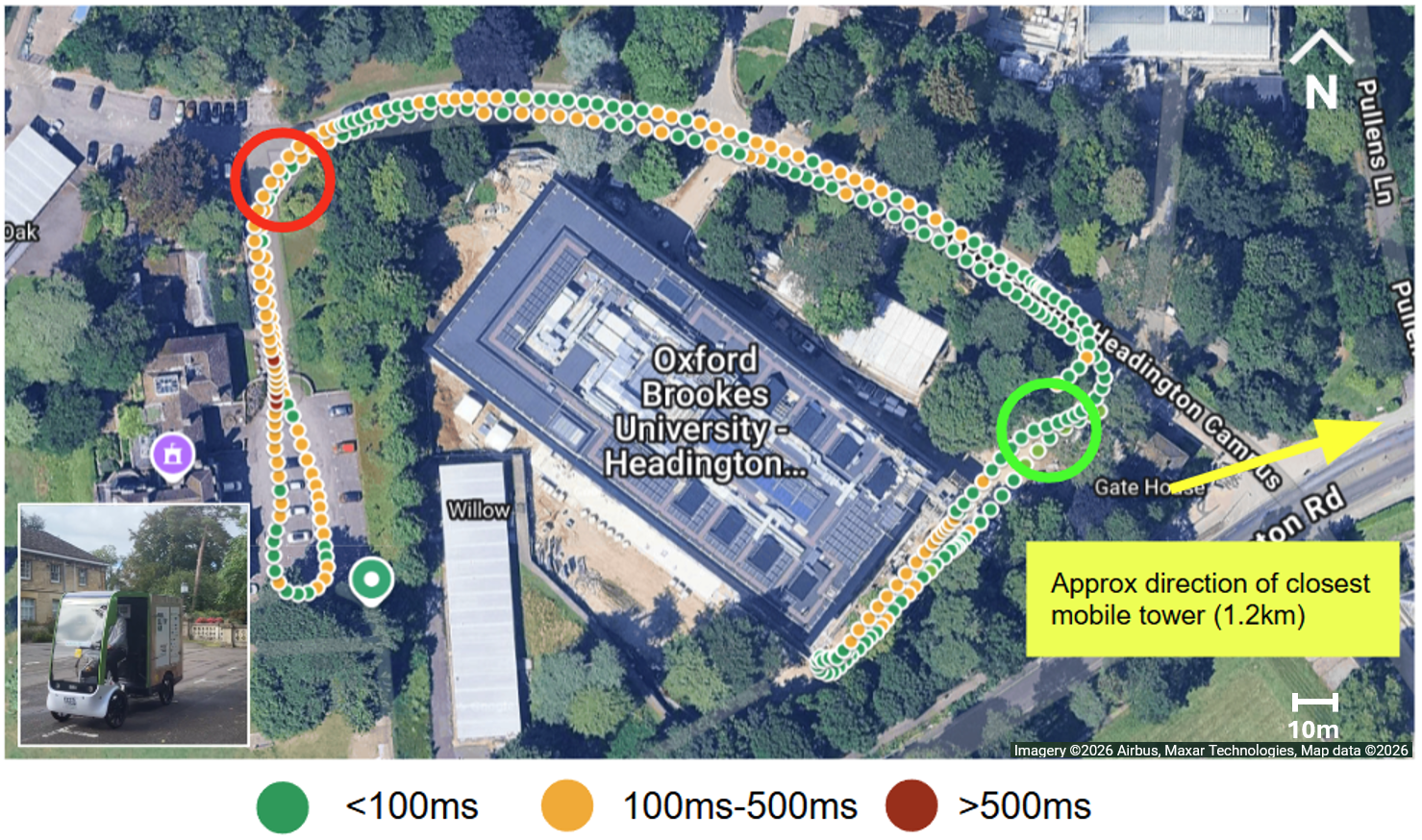}
    \caption{Variation in network latency captured while driving the ROV on campus streaming data with a rate of 1000 kbit/s upload. Sample areas for poor signal strength (circled red) and improved signal strength (circled green) are marked for further evaluation. Note that areas of the route occluded from the mobile mast by the building exhibit higher latency. Imagery ©2026 Airbus, Maxar Technologies, Map data ©2026 Google.}
    \label{fig:mapLatency}
\end{figure}

To evaluate the constraints and performance of the technique, a number of experiments were carried out. These included the following: Quantifying network performance within the test area; Evaluating the Glass-to-Glass (G2G) performance of the SEG-JPEG technique compared with h.264 compression; Evaluating imagery quality. All remote operation testing was conducted at Oxford Brookes University's Headington Hill Campus with a safety driver on-board using public mobile phone networks. The approximately 0.8 km long route (Fig.~\ref{fig:mapLatency}) was selected to include areas of varying connectivity to reflect real world conditions across a larger urban setting. Due to interrupted 5G connectivity, testing was conducted using the 4G network using a single provider. To evaluate how the network coverage affects latency, a continuous stream of UDP packages contributing to a data rate of 1 Mbit/s was transmitted via the public mobile network from the ROV to the remote operation station, connected to a wired local network.

\section{Results And Evaluation}

\subsection{Network Saturation And Latency}
  The West side of the route exhibited substantial latency due to a large, elevated campus building in the line-of-sight between the vehicle and the network provider's nearest mobile mast (approximately 1.2 km to the East). With a 1 Mbit/s data rate, the median network latency across the entire test was evaluated to be 82 ms with a standard deviation of 32 ms.

\begin{figure} [ht]
    \centering
    \includegraphics[width=1\linewidth]{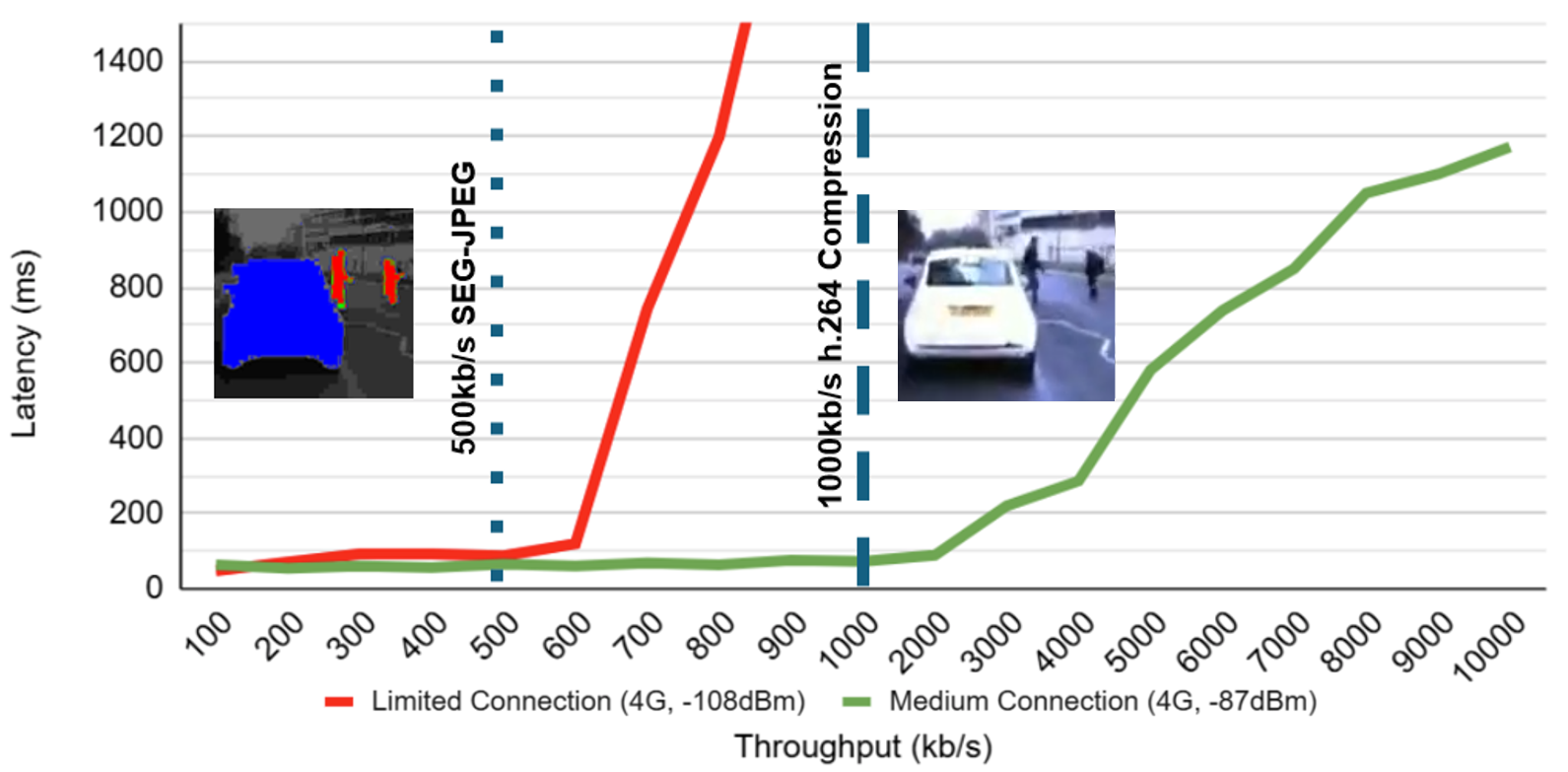}
    \caption{Latency by mobile network throughput, showing how increasing the data rate of streamed imagery increases latency on a limited 4G connection.}
    \label{fig:LatencyByMobileNetThru}
\end{figure}

Areas of very poor signal strength (-108 dBm) and poor signal strength (-87 dBm) were reserved to evaluate the maximum low latency throughput in more detail. The data rate was increased by 100 kbit/s at regular intervals until the network latency exceeded 5 s. The saturation point is considered to be the point when the network latency exceeds 100 ms, since the latency increases substantially past this threshold. At a signal strength of -108 dBm the network saturation point was observed to be reached at a data rate of 600 kbit/s. With a slightly improved signal strength of -87 dBm the saturation point increased significantly to 2000 kbit/s (Fig.~\ref{fig:LatencyByMobileNetThru}).

As throughput increases, the network reaches a saturation point where the latency increases significantly, providing an upper limit beyond which the network becomes unsuitable for use for remote operation. It can be seen that the 1000 kbit/s data rate required for a traditional h.264 stream is not suitable for streaming on a limited connection, and lies close to the upper limit for a medium quality connection. SEG-JPEG, on the other hand is able to maintain stable operation at a network throughput of just 500 kbit/s, enabling robust communication and avoiding latency spikes, thus facilitating safe operation of the vehicle in poor signal areas.

\subsection{Glass-to-Glass Performance}
\begin{figure} [ht]
    \centering
    \includegraphics[width=0.9\linewidth]{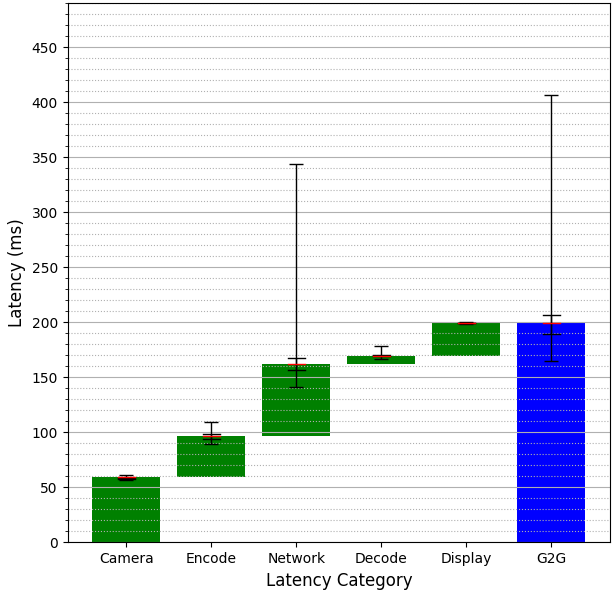}
    \caption{G2G latency distribution for SEG-JPEG imagery sent over a poor quality 4G mobile network connection (10Hz, 500kbit/s). The network latency is the largest contributor to overall (G2G) latency variation.}
    \label{fig:G2GLatency}
\end{figure}

Using a 4G connection, SEG-JPEG achieved a median G2G latency of 198 ms with a throughput of 500 kbit/s at a frame rate of 10 Hz (Fig.~\ref{fig:G2GLatency}). h.264 achieved a similar median G2G latency of 195 ms at the same throughput but with a frame rate of 30 Hz. Interestingly, decreasing the frame rate to 10 Hz increased the G2G latency to above 500 ms when using h.264 compression. It was observed that the h.264 imagery became unusable with the vehicle in motion due to frame distortions such as tearing and macroblocking (Fig.~\ref{fig:DistortedH264}). 

\subsection{Imagery Quality}

\begin{figure} [ht]
    \centering
    \includegraphics[width=1\linewidth]{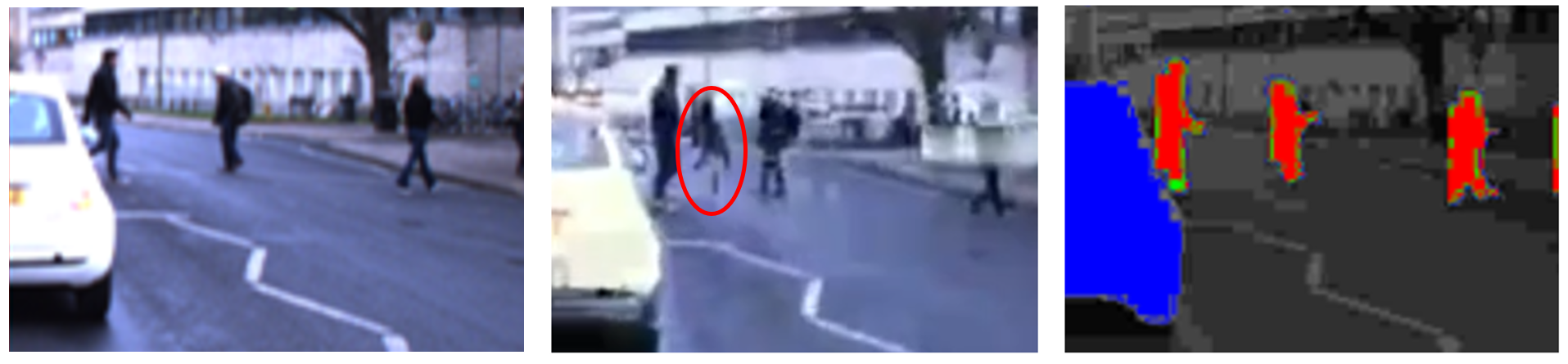}
    \caption{Original uncompressed imagery (left), distorted 500 kbit/s h.264 imagery exhibiting ghosting where remnants of a moving pedestrian from a previous frame are visible (centre), 500 kbit/s SEG-JPEG imagery (right)\protect\footnotemark[1].}
    \label{fig:SideBySide}
\end{figure}

Qualitative evaluation of image clarity makes it immediately apparent that the imagery compressed using h.264 experiences distortion such as ghosting when the network throughput is insufficient (Fig.~\ref{fig:SideBySide}, right). It can be seen that even though the h.264 imagery appears to better represent the original uncompressed imagery, the parts of the imagery which are most important for the RO, such as the pedestrians crossing the road, are obscured by the distortion resulting in potential misdetections which could significantly affect situational awareness. Unlike simply adding bounding boxes to highlight pedestrian locations, SEG-JPEG allows the operator to see and understand nuances within the image \textit{e.g.} the motion of a pedestrians arms and legs - aiding the operator in understanding the behaviour of, and predicting future trajectories of other road users; crucial to safe operation of the vehicle \cite{teetiASTRASceneawareTRAnsformerbased2025}.

\subsection{Results Summary}

\textcolor{black}{It was established that the low-latency throughput capacity of a public mobile network is limited by signal strength. Limiting the imagery data rate to 500 kbit/s enabled sub-200 ms G2G latency even on a very limited (-108 dBm) 4G network connection. Compression associated artifacts experienced using rate-limited h.264 compression could be avoided by highlighting important road users using SEG-JPEG.}

\section{Discussion}
While many recent studies have presented remote driving concepts 
\cite{fezeuTeleoperatingAutonomousVehicles2026},\cite{bellangerChallengesRemoteDriving2024},  \cite{kerblTUMTeleoperationOpen2025}, \cite{testouri5GEnabledTeleoperatedDriving2025}, they typically require high quality mobile connectivity with a network throughput of at least 20 Mbit/s to transmit the imagery needed to safely remotely operate a road vehicle. This comparatively large throughput requirement restricts remote operation deployments to geographically limited areas where network quality exceeds nation-wide averages \cite{ofcomConnectedNationsUK2025}. Typically, such requirements dictate the use of capital-intensive private 5G networks. As the number of remote operation deployments increase over time, an increasing number of vehicles will have to share the same network cell, potentially blocking future developments. Lucas-Estan \textit{et al.} \cite{lucas-estanSupportTeleoperatedDriving2023} found that a density of just 10 vehicles/km\textsuperscript{2} remotely operated using conventional streaming methods would exceed the upload capacity of an idealised 5G network. Even with still-higher-bandwidth 6G and beyond communications technologies, inevitable capacity limits will mean that the types of bandwidth-conserving, QoS-robust edge compression techniques developed here will have continuing value, permitting wide-scale deployments.

While evaluating remote operation on real-world, public mobile networks, 4G was found to provide more robust streaming, due to the increased stability of the connection. The 5G signal loss experienced is likely caused by buildings obscuring the path to the nearest mobile tower, compounded by penetration losses caused by the use of high frequency millimetre-wave signals \cite{batistaMethodologyEstimatingRadiofrequency2025}. The lower frequencies typically used by the 4G mobile network can result in less penetration loss \cite{bagheriEnhancing5GPropagation2024}, which can increase coverage, particularly in built-up urban areas, resulting in fewer communication drop-outs.

It was observed that sub-200 ms G2G latency can be achieved on reduced quality mobile networks, provided that the data rate of imagery is limited to less than the maximum throughput of the connection. A saturation point was established, above which the latency of the transmitted imagery continued to increase at a rate proportional to the increase in the data rate in relation to the throughput, similar to the cumulative delay effect observed by Fezeu \textit{et al.} \cite{fezeuTeleoperatingAutonomousVehicles2026}.

\textcolor{black}{Under extremely poor network conditions,} greyscale JPEG imagery was observed to exhibit significantly less compression-associated distortion compared with low-latency optimised h.264 imagery. \textcolor{black}{The h.264 codec is based on inter-frame compression, where each frame observed by the RO is a reconstruction based upon a series of motion vector updates on top of occasional key-frames \cite{wiegandOverviewH264AVC2003}. As network quality reduces, frame updates are more likely to arrive with varying delay and/or missing information, resulting in an accumulation of frame mis-matches on the receiving end - causing distortion such as tearing and/or ghosting (Fig.~\ref{fig:SideBySide}, centre). SEG-JPEG uses frame-by-frame (intra-frame) compression, where each image observed by the RO is an independent frame, thus avoiding cumulative distortion artifacts (Fig. ~\ref{fig:SideBySide}, right).} By encoding salient information into greyscale imagery, SEG-JPEG represents an application of semantic communication which, rather than trying to reconstruct a visually accurate image \cite{huangSemanticCommunicationsDeep2023} at high computational cost, transmits a low quality image containing important road user information which can be easily interpreted by a RO. 

It is envisaged that SEG-JPEG could operate as part of a smart, adaptive streaming system - where \textit{e.g.} high quality video is sent when bandwidth is available, and SEG-JPEG replaces poor-quality, high latency or visually inaccurate streaming as network quality worsens; facilitating robust, low latency communication between vehicle and RO on real-world, variable quality mobile networks, and thus significantly extending the Operational Design Domain of the vehicle.

\section{Conclusion}
Leading remote operation implementations rely heavily upon low latency, high throughput network conditions. This represents an obstacle to wider remote operation deployments, as public mobile networks often experience considerable throughput constraints; resulting in high latencies, and/or a loss of visual information which can render the remote operator unable to safely control or command the vehicle. SEG-JPEG provides a solution to this challenge - making use of object detection and semantic segmentation, coupled with a method of augmenting highly-compressed greyscale imagery with representations of the detected road users; in a format that can be recoloured at the receiving end - thus enhancing the situational awareness of the remote operator. This technique reduces the minimum throughput required per stream to 500 kbit/s - a reduction of up to 50\% compared to the leading technique (h.264), resulting in a median glass-to-glass latency of 200 ms over a public 4G mobile network in sub -100 dBm signal strength areas. This facilitates robust communication and safe remote operation of automated vehicles via public 4G mobile networks, significantly extending the operational design domain and thus providing the potential to accelerate nationwide roll-outs of automated vehicles.

\bibliographystyle{unsrturl}
\bibliography{references}
\end{document}